\documentclass[letterpaper]{article} % DO NOT CHANGE THIS
\usepackage{aaai2026}  % DO NOT CHANGE THIS
\usepackage{times}  % DO NOT CHANGE THIS
\usepackage{helvet}  % DO NOT CHANGE THIS
\usepackage{courier}  % DO NOT CHANGE THIS
\usepackage[hyphens]{url}  % DO NOT CHANGE THIS
\usepackage{graphicx} % DO NOT CHANGE THIS
\usepackage{natbib}  % DO NOT CHANGE THIS AND DO NOT ADD ANY OPTIONS TO IT
\usepackage{caption} % DO NOT CHANGE THIS AND DO NOT ADD ANY OPTIONS TO IT
\usepackage{algorithm}
\usepackage{algorithmic}

\usepackage{newfloat}
\usepackage{listings}
\DeclareCaptionStyle{ruled}{labelfont=normalfont,labelsep=colon,strut=off} % DO NOT CHANGE THIS
\floatstyle{ruled}
\newfloat{listing}{tb}{lst}{}
\floatname{listing}{Listing}
\usepackage{amsmath}
\usepackage{amsfonts}
\usepackage{booktabs}
\usepackage{multirow}

\title{Scaling LLM Speculative Decoding:\\ Non-Autoregressive Forecasting in Large-Batch Scenarios}
\author {
    Luohe Shi\textsuperscript{\rm 1},
    Zuchao Li\textsuperscript{\rm 2}\thanks{Corresponding Author},
    Lefei Zhang\textsuperscript{\rm 1},
    Baoyuan Qi\textsuperscript{\rm 3},
    Guoming Liu\textsuperscript{\rm 3},
    Hai Zhao\textsuperscript{\rm 4}
}
\affiliations {
    \textsuperscript{\rm 1}School of Computer Science, Wuhan University  \\ 
    \textsuperscript{\rm 2}School of Artificial Intelligence, Wuhan University  \\
    \textsuperscript{\rm 3}Xiaomi  \\
    \textsuperscript{\rm 4}School of Computer Science, Shanghai Jiao Tong University \\
    \{shiuohe, zcli-charlie, zhanglefei\}@whu.edu.cn  \\ 
    \{qibaoyuan, liuguomin\}@xiaomi.com  \\
    zhaohai@cs.sjtu.edu.cn
}

\begin{document}

\maketitle

\begin{abstract}
Speculative decoding accelerates LLM inference by utilizing otherwise idle computational resources during memory-to-chip data transfer. Current speculative decoding methods typically assume a considerable amount of available computing power, then generate a complex and massive draft tree using a small autoregressive language model to improve overall prediction accuracy. However, methods like batching have been widely applied in mainstream model inference systems as a superior alternative to speculative decoding, as they compress the available idle computing power. Therefore, performing speculative decoding with low verification resources and low scheduling costs has become an important research problem. We believe that more capable models that allow for parallel generation on draft sequences are what we truly need. Recognizing the fundamental nature of draft models to only generate sequences of limited length, we propose SpecFormer, a novel architecture that integrates unidirectional and bidirectional attention mechanisms. SpecFormer combines the autoregressive model’s ability to extract information from the entire input sequence with the parallel generation benefits of non-autoregressive models. This design eliminates the reliance on large prefix trees and achieves consistent acceleration, even in large-batch scenarios. Through lossless speculative decoding experiments across models of various scales, we demonstrate that SpecFormer sets a new standard for scaling LLM inference with lower training demands and reduced computational costs.
\end{abstract}

\begin{links}
    \link{Code}{hhttps://github.com/ShiLuohe/SpecFormer}
\end{links}


\section{Introduction}

Large language models (LLMs) that utilizes Transformer Decoders have rapidly become the industry standard in recent years, owing to their favorable properties such as scalability in training and lossless handling of long-context dependencies~\citep{DBLP:journals/corr/abs-2303-08774}. Nevertheless, these models continue to follow the conventional sequence-to-sequence generation paradigm: autoregressive decoding. Autoregressive decoding refers to the process where tokens are generated one at a time; each newly generated token is fed back into the model as input for the next step, alongside the existing context, to perform another forward pass. This paradigm offers several notable advantages. With only a causal mask, the attention mechanism can be readily adapted for generation tasks, making training straightforward. It also allows for the generation of virtually unlimited-length outputs and enables acceleration through state caching for repeated input prefixes~\citep{DBLP:journals/corr/abs-2407-18003}, and some corresponding acceleration opportunities~\citep{guo-etal-2025-tom, yang-etal-2025-xquant, DBLP:journals/corr/abs-2508-02751, zhao-etal-2025-dac, tang-etal-2025-spindlekv}.

However, during inference, generating one token at a time results in low arithmetic intensity (AI, \citealp{DBLP:journals/cacm/WilliamsWP09}). Once a datum fetched from memory into the chip, it contributes to only a few operations. In contrast, chips can perform hundreds of operations in the time, leading to substantial underutilization of compute resources. On the infrastructure side, techniques such as prefill-decoding (PD) separation~\citep{298687} and continuous batching~\citep{280922, 10.1145/3600006.3613165} have been introduced to improve overall compute utilization and user experience. Nonetheless, under constraints imposed by service-level objects (SLOs, \citealp{DBLP:journals/corr/abs-2410-14257}), these techniques often fall short of leveraging the full computational potential. Increasing AI, the ratio of computation to data transfer, is the fundamental approach to enhancing hardware efficiency during generation.

Speculative decoding (SD, \citealp{xia-etal-2024-unlocking}) is one of the most effective approaches for improving AI. Its core idea is to generate multiple tokens per pass of the large model. The process consists of three main steps~\citep{NEURIPS2018_c4127b91}: 
1. {\bf Multi-token generation}: Based on information from the previous forward pass, the model samples multiple draft tokens.
2. {\bf Multi-token verification}: The model evaluates all draft tokens simultaneously to determine whether each one aligns with its own top prediction, while also extracting and storing information for the next round of multi-token generation.
3. {\bf Multi-token acceptance}: The model decides whether to accept the draft tokens based on the verification results and accordingly updates the contextual information.

% Since multiple tokens are generated in one forward pass, speculative decoding is also referred to as multi-token prediction (MTP). 
{\bf Multi-token generation} is the most critical component of SD, as the acceptance rate of the sampled drafts directly determines how effectively computational resources are utilized. {\bf In this work, we focus specifically on lossless SD}, which adheres to two strict conditions: 1. Only draft tokens that exactly match the outputs of the large model are accepted. 2. The LLM itself must remain unmodified.
These constraints make a purely acceleration-oriented SD, ensuring strict mathematical equivalence with the original model outputs. 
% While relaxing accuracy requirements can potentially yield further speedup, the resulting performance degradation is often difficult to quantify or control, and may unnecessarily complicate the problem. Moreover, empirical observations suggest that algorithms performing well under the lossless SD setting tend to also exhibit strong performance under lossy conditions. 

A key observation about SD is that it does not reduce (usually increases significantly) the total amount of computation. The acceleration arises from repurposing compute capacity that would otherwise be idle while waiting for data transfer. In other words, every SD-based method has a theoretical upper bound on speedup, corresponding to the full utilization of previously wasted compute. It is important to note that continuous batching is also a method for reducing idle compute. Consequently, in batched settings where unused compute capacity is already diminished, speculative decoding methods face a stricter efficiency requirement.

Current SD methods can be broadly categorized into autoregressive and non-autoregressive approaches~\citep{DBLP:journals/corr/abs-2502-19732}. 
They all generate draft tokens in time that scales linearly with the number of draft tokens, whether through autoregression or by accessing different parameters.
% The former 
% typically employs a smaller auxiliary causal model: during generation, the small model accesses partial states from the large model and uses autoregressive decoding to rapidly generate multiple subsequent tokens—benefiting from its smaller size and lower computational cost. The latter, in contrast, utilizes non-autoregressive techniques by storing multiple sets of position-specific parameters. These parameters are used to directly generate draft tokens from the large model’s internal states, with each parameter set responsible for predicting a draft token at a specific future position. 
% In both paradigms, these models are often combined with prefix trees, where instead of sampling a single best draft sequence, multiple suboptimal candidates are sampled in parallel~\citep{DBLP:conf/emnlp/LiW0024}. These are merged into a prefix tree and collectively verified by the large model. 

\begin{figure}[ht]
    \centering
    \includegraphics[width=\linewidth]{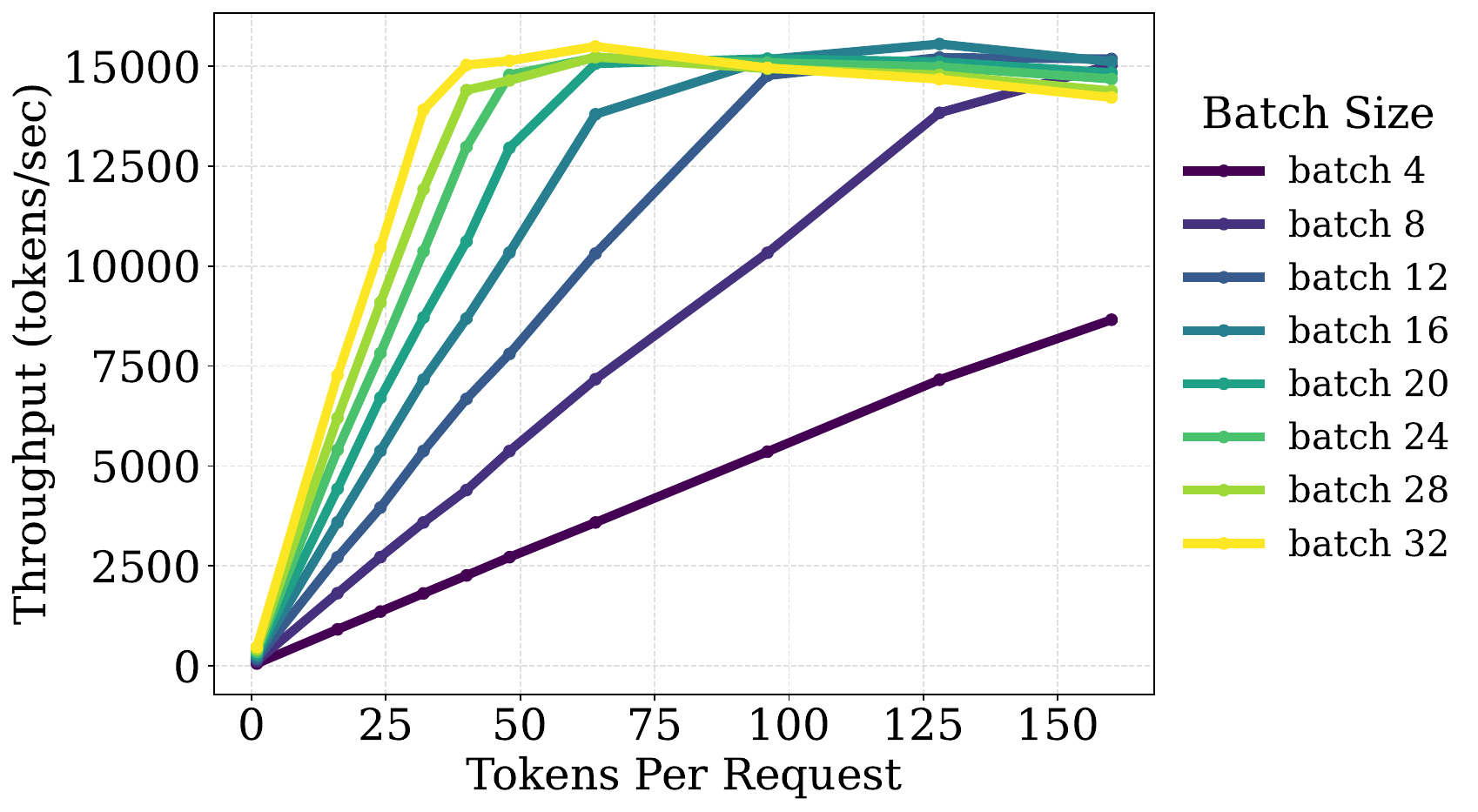}
    \caption{How batch size affects the max draft size.}
    \label{fig:tpt}
\end{figure}

% However, a key challenge arises in batched inference settings, where the residual compute capacity available for speculative decoding is significantly reduced. This severely limits the size of the prefix tree, often degenerating it into a single linear sequence, thereby limiting the predictive accuracy that could otherwise be gained from broader exploration. Moreover, the auxiliary model's parameters are either position-dependent or autoregressive, require repeated memory access for sequential decoding. In both cases, scaling up the auxiliary model to improve prediction quality becomes difficult, as it incurs substantial additional cost or inefficiency.
However, we've noticed a significant issue with current methods: their incompatibility with large batch sizes. A larger batch size means each parameter experiences higher computational intensity, as it's read and reused multiple times. This, in turn, reduces the computational cycles available for draft tokens. In other words, under batch processing conditions, there are twice as many draft tokens to process in less available time. This point is illustrated in Figure~\ref{fig:tpt}, where we show that when processing the same number of tokens while varying the batch size, the speed at which peak computational power is reached accelerates with increasing batch size.

The number of draft tokens we can use is less than what's needed to reach peak performance. This means that in a batched environment, the usable draft tree size is rapidly compressed. Given that online batching has been widely adopted by mainstream inference frameworks, current speculative decoding methods must adapt to scarcer resources. This implies they can no longer rely on traditional, massive draft trees, but instead need to focus on higher accuracy drafts themselves. Mainstream models face a significant challenge in this regard: they struggle to efficiently scale to larger sizes to displace capabilities. This is due to their excessive position-dependent parameters—whether these are parameters that need to be repeatedly accessed for each position in AR methods, or parameters separately allocated for each position in non-AR methods. The problem with position-dependent parameters is that scaling up exponentially increases additional costs because multiple tokens are required in a sequence. This means these methods not only inherently use more computational power but are also highly sensitive to changes in available compute.

Therefore, we aim to improve the performance of SD under low draft token budgets, by directly enhancing the capability of the draft generation model. This enables SD to be effectively applied in batched inference settings. To avoid fine-tuning the original LLM, the draft model must receive sufficiently rich input information. To this end, we employ a context causal attention to extract contextual information from the hidden states of the input sequence. We observe that in traditional approaches, the parameters used for draft generation are position-dependent, i.e., generating each position in the draft sequence typically requires accessing a large number of parameters tied to that specific position. Instead, we seek a prediction mechanism in which the majority of parameters are position-independent, while retaining only a limited amount of positional information. Furthermore, we identify a key distinction between draft generation in SD and open-ended generation in LLMs: SD only requires a small number of future tokens, rather than unbounded generation. Motivated by this, we adopt a Draft Bi-directional Attention architecture for draft token generation. This forms the basis of our proposed SpecFormer architecture.

More specifically, we enable the model to efficiently extract information from both greater depth and breadth simultaneously through multi-level hidden state feature fusion combined with a causal masked attention mechanism. Subsequently, by assigning a specific set of matrix multiplication weights to each position, we can effectively inject positional information to obtain the initial state for each subsequent token. Finally, a standard Encoder layer is used to parallelize the fine-grained generation of tokens at each subsequent position.

We evaluate our proposed method on models of approximately 4B, 7B, and 14B parameters~\citep{qwen2.5}, conducting both theoretical and real-world experiments. In the theoretical experiments, we constrain the number of draft tokens to simulate varying levels of redundant computational capacity and draft model cost. Under these conditions, we measure the average accepted token length across different methods to assess their efficiency. In the real-world experiments, we evaluate the acceleration ratio of our method under different batch size settings using dialogue datasets and standard benchmarks, demonstrating its effectiveness in practical deployment scenarios.

\section{Background and Related Works}

\subsection{Non-autoregressive SD Approaches}

Non-autoregressive methods refer to SD algorithms in which the draft tokens are generated without causal dependencies among them. The most common examples include Multi-Token Prediction (MTP, \citealp{DBLP:conf/icml/GloeckleIRLS24}) and Medusa~\citep{DBLP:conf/icml/CaiLGPLCD24}. These approaches share a common principle: leveraging the last hidden state (LHS) of the LLM, originally used for predicting the next token, to predict multiple future tokens simultaneously. Medusa trains a separate MLP layer for each target position, projecting the LHS into a new token space, which is then fed into the LM\_Head to generate the corresponding draft token. In contrast, MTP designs multiple LM\_Heads, each dedicated to generating the draft token at a specific future position. Positional-sharing parameters are typically less while Positional-specific ones remains fairly many for these methods. These methods often suffer from limited predictive capacity due to their inability to access information from the entire sequence, and they typically require fine-tuning the entire model.

\subsection{Autoregressive SD Approaches}

Autoregressive methods employ a smaller sequence model to generate future tokens autoregressively based on the input sequence. Autoregressive decoding can operate at three levels:
\begin{enumerate}
    \item {\bf Token level}: These methods use a standalone small language model (SLM) to generate future tokens autoregressively. The SLM typically shares the same vocabulary as the LLM. It receives the input tokens from the LLM, samples several future tokens autoregressively, and then passes them to the LLM for validation. A key advantage is that, if a suitable SLM exists, no additional training is required. However, such models are difficult to obtain, and the approach introduces significant KV cache overhead. A representative method is BiLD~\citep{NEURIPS2023_7b97adea} decoding~\citep{DBLP:conf/emnlp/Xia0WCWS23, DBLP:journals/corr/abs-2405-19715, DBLP:conf/iclr/ZhouLRMRKKA24, bachmann2025judge}.
    \item {\bf LHS level}: These methods perform autoregressive decoding over LHS representations. A small model consumes the LHS output from the LLM, predicts the next LHS, and recursively feeds it into itself. The resulting LHSs are then converted to token predictions and validated by the LLM. The small model is typically a decoder layer and requires additional training, but since the LLM itself is not modified, the training cost in both time and memory is significantly lower than fine-tuning. The primary limitation lies in the difficulty of aligning the small model to the LHS space, which can impair its performance. Representative methods include EAGLE~\citep{li2025eaglespeculativesamplingrequires}, HASS~\citep{zhang2025learning}, Deepseek-V3 MTP~\citep{DBLP:journals/corr/abs-2412-19437}, etc.\citep{Gao_Xie_Xiang_Ji_2025, chen-etal-2025-faster}
    \item {\bf Independent representation} : These methods construct a separate latent space by combining the LLM's LHS with auxiliary information, and perform autoregressive decoding in this space. A notable example is EAGLE-3~\citep{DBLP:journals/corr/abs-2503-01840}.
\end{enumerate}
A common challenge across autoregressive decoding models is that the repeated invocation of the small model means that, even with identical content, its parameters remain position-dependent, leading to higher computational costs. Furthermore, due to the limited capacity of the small model, these methods often require a very wide prefix tree to explore multiple hypotheses in parallel, in order to achieve acceptable prediction accuracy.

\section{Methods}

\subsection{From Arithmetic Intensity to SD Evaluation}

Arithmetic intensity (AI) is defined as the ratio between the number of required floating-point operations and the number of bytes of data that must be read. For a model with $M$ parameters operating in half-precision, the arithmetic intensity $AI_m$ is given in Equation~\ref{eq:ai}.
\begin{equation}
\label{eq:ai}
\begin{aligned}
    AI_m & = \frac{\mathrm{Model\ FLOPS}}{\mathrm{Memory\ I/O}} \\
    & = \frac{2 \cdot M}{\mathrm{bytes(\mathbf{bf16})}\cdot M}
    = 1
\end{aligned}
\end{equation}
For an acceleration chip, we can estimate the ideal arithmetic intensity $AI_c$ required to fully utilize its compute capacity by examining the ratio of its peak FLOPs to memory bandwidth (typically DDR, GDDR, or HBM). For Tesla A100-80G, the $AI_c$ as shown in Equation~\ref{eq:aih}.
\begin{equation}
\label{eq:aih}
\begin{aligned}
    AI_c(\mathrm{A100}) & = \frac{\mathrm{Peak\ FLOPS}\ \mathbf{bf16}}{\mathrm{Memory\ Bandwidth}} \\
    & = \frac{311.84\ \mathrm{TFLOPS/s}}{2.04\ \mathrm{TB/s}} = 152.86
\end{aligned}
\end{equation}
We define the redundancy ratio $\rho$ as the ratio $AI_c/AI_m$, which represents both the ideal batch size and the theoretical upper bound of speedup achievable through batching effects. It should be noted that due to practical factors such as scheduling overhead, $\rho$ does not reflect actual performance precisely, but it provides a useful baseline for system-level analysis. We prefer smaller values of $\rho$, as a lower $\rho$ indicates less wasted compute, with $\rho=1$ representing the ideal case where no redundancy remains. 

Previous work on SD has typically focused on average accepted token length, i.e., the average number of tokens accepted per invocation of the LLM. However, we argue that this metric is overly coarse-grained: it obscures the underlying total computational cost, making it difficult to adapt methods to new deployment scenarios. We contend that a better criterion for evaluating an SD method is to examine its performance under a fixed draft token budget.

For a given SD algorithm, suppose it increases the total computation by a factor of $p$, generates $k$ draft tokens per step, and among them, an average of $a$ tokens are accepted. Then, the effective AI gain relative to standard LLM decoding which we want to maximize, denoted as $r_1$, and the on-chip AI gain, denoted as $r_2$, can be derived as a function of $\rho$ in Equation~\ref{eq:goal}, with $bs$ representing the batch size.
\begin{equation}
\label{eq:goal}
    \max r_1 = \frac{a}{p}AI_m,\quad\mathbf{s.t.}\ r_2 = k \leq \frac{\rho}{bs}
\end{equation}
Moreover, if the SD model requires $m_p$ parameters to generate draft token of each position, and $m_s$ parameters that shares within all positions, with the draft sequence length $l_d$, the $p$ is given in Equation~\ref{eq:comp}.
\begin{equation}
\label{eq:comp}
    p= 1 + \frac{m_s+l_d\cdot m_p}{M}
\end{equation}
Finally, we define an optimization coefficient $\kappa$ in Equation~\ref{eq:met}, which captures the model’s ability to accelerate under constrained resources. We aim to maximize $\kappa$, or increase the draft token acceptance rate while minimizing the computational overhead of the draft model. In prior work, $k$ was often either ignored or fixed to a relatively large constant, owing to the availability of abundant redundant compute. However, as the batch size increases, the available redundant compute rapidly diminishes, making $k$ a critical factor that significantly impacts performance.
\begin{equation}
\label{eq:met}
    \kappa = \frac{a \cdot l_d}{k} 
\end{equation}

\begin{figure*}[ht]
\centering
\begin{minipage}{0.59\textwidth}
    \centering
    \includegraphics[width=1\linewidth]{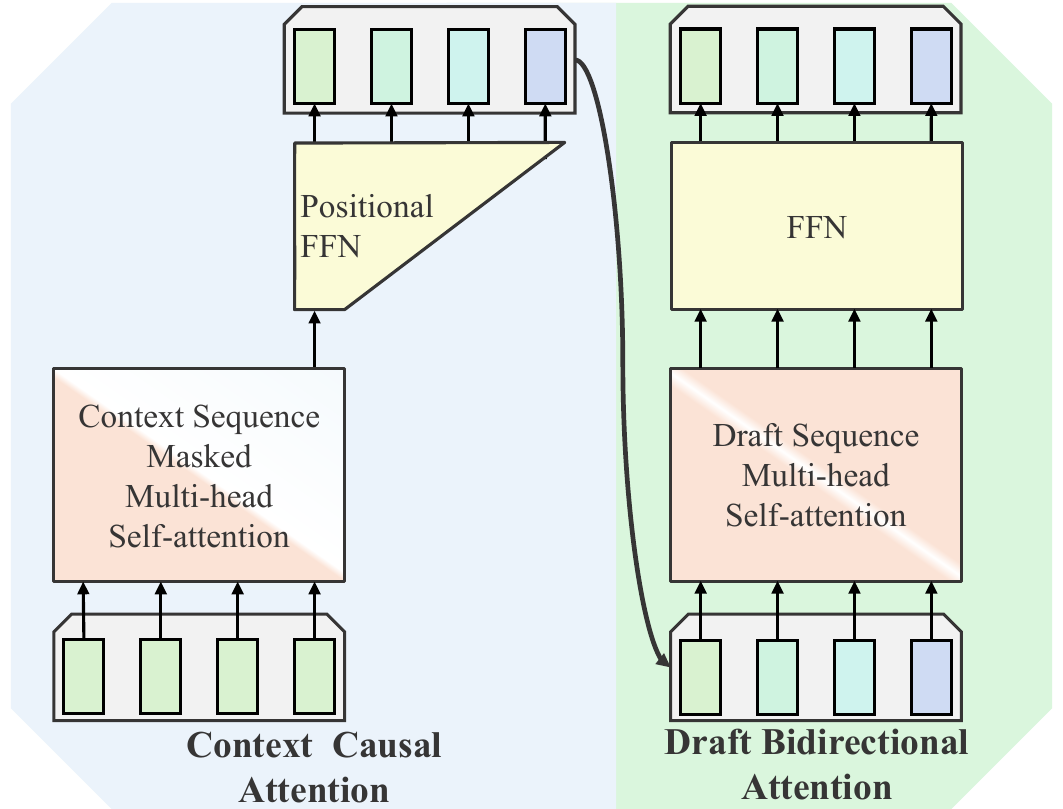}
    \caption{An overview of proposed SpecFormer speculative decoding method.}
    \label{fig:overv}
\end{minipage}
\hfill
\begin{minipage}{0.36\textwidth}
    \centering
    \includegraphics[width=1\linewidth]{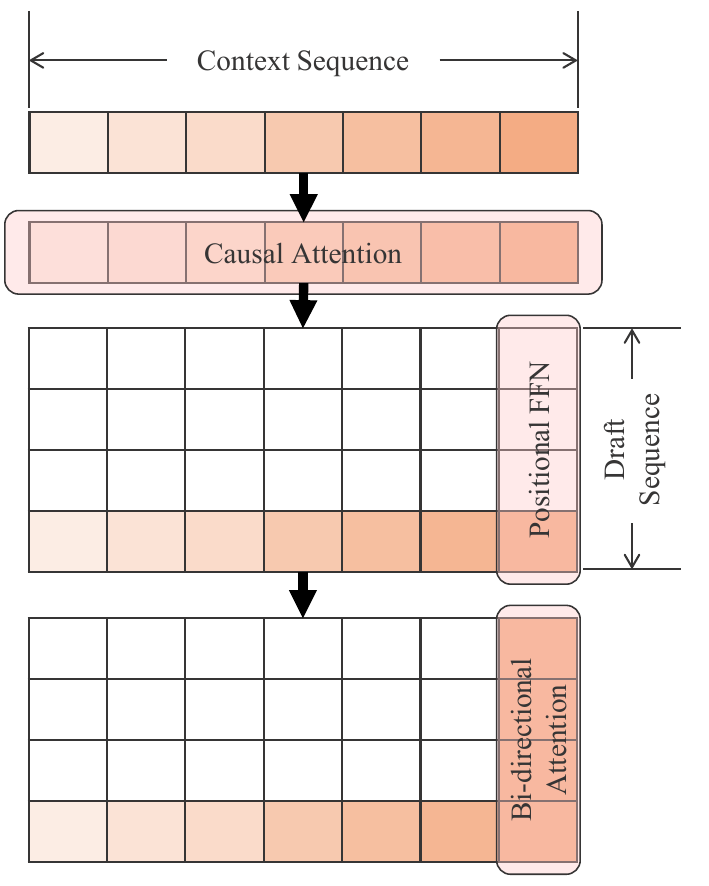}
    \caption{A depiction of uni and bi-directional attention.}
    \label{fig:biatt}
\end{minipage}
\end{figure*}

\subsection{General Notations}

\newcommand\Corp{\mathcal{C}}
\newcommand\Abs[1]{\left| #1 \right|}

We define $\Corp$ as our training corpus with $\Abs{\Corp}$ entries. An entry $c \in \Corp$ is a list a tokens $x_1x_2\dots x_{\Abs{c}}$. The training goal of next-token prediction for LLM pretraining is to find the $\theta_{\mathcal{LM}}$ that minimize the cross entropy loss, given in Equation~\ref{eq:ntp}. For an SD module with parameters $\theta_{\mathcal{SD}}$ and maximum drafting length $l_d$, the optimizing goal is given in Equation~\ref{eq:mtp}.
\begin{subequations}
    \begin{equation}
    \label{eq:ntp}
        \mathrm{arg}\min_{\theta_{\mathcal{LM}}}
        \sum_{c\in\Corp}
            \sum_{i=2}^{\Abs{c}}
                \frac{
                    -\log P_{\theta_{\mathcal{LM}}}\left(
                        x_i\ |\ x_1\dots x_{i-1}
                    \right)
                }{\Abs{\Corp}\cdot\Abs{c}}
    \end{equation}
    \begin{equation}
    \label{eq:mtp}
    \begin{aligned}
        & \mathrm{arg}\min_{\theta_{\mathcal{SD}}} \\
        & \sum_{c\in\Corp}
            \sum_{j=2}^{l_d + 1}
                \sum_{i=1+j}^{\Abs{c}} \!\!
                \frac{
                    -\log P_{\left(\theta_{\mathcal{LM}}, \theta_{\mathcal{SD}}\right)}\left(
                        x_i\ |\ x_1\dots x_{i-j}
                    \right)
                }{\Abs{\Corp}\cdot\Abs{c}\cdot l_d}
    \end{aligned}
    \end{equation}
\end{subequations}

We further denote $L$ as the layer count of the base LLM and $d_h$ as the hidden size. Hidden states $\mathrm{HS}\in\mathbb{R}^{(L+1)\times\Abs{c}\times{d_h}}$ are the states that traversing between layers, where $\mathrm{HS}[i]$ represents the $i$-th layer of HS. Specifically, $\mathrm{HS}[0]=\text{Embedding}(c)$ and $\mathrm{LHS}=\mathrm{HS}[L]$. 

Finally, we define Equation~\ref{eq:com} to simplify the description of pre-norm residual connected units.
\begin{equation}
\label{eq:com}
    (\mathrm{Ops} \cdot \mathrm{Norm} + \mathbb{I})(X)  \iff
        \mathrm{Ops}(\mathrm{Norm}(X)) + X
\end{equation}

\subsection{SpecFormer}
% \begin{figure}[ht]
%     \centering
%     \includegraphics[width=1\linewidth]{figures/FigMain.pdf}
%     \caption{An overview of proposed SpecFormer speculative decoding method.}
%     \label{fig:overv}
% \end{figure}
% \begin{figure}[ht]
%     \centering
%     \includegraphics[width=0.8\linewidth]{figures/FigBiAttn.pdf}
%     \caption{A depiction of uni and bi-directional attention.}
%     \label{fig:biatt}
% \end{figure}

Our proposed SpecFormer comprises a Context Causal Attention and a Draft Bi-directional Attention, depicted in Figure~\ref{fig:overv}, incorporating both unidirectional and bidirectional attention along two dimensions, as shown in Figure~\ref{fig:biatt}.

\subsubsection{Context Causal Attention}

The Context Causal Attention module consists of three components: Hook and Downsampler, Causal Attention, and a Positional Feedforward Network (Positional FFN). Each component takes inputs passed through root-mean-square normalization (RMS Norm), with the Downsampler employing Grouped RMS Norm and the Causal Attention module utilizing a residual connection.

The hook module extracts information from the HS, following the approach introduced by \citeauthor{DBLP:journals/corr/abs-2503-01840}. Specifically, we select four layers: $\mathrm{HS}[0]$, $\mathrm{HS}[L/2]$, $\mathrm{HS}[L-1]$, and $\mathrm{HS}[L]$, and concatenate them to form a tensor $I\in \mathbb{R}^{bs\times\Abs{c}\times4\times dh}$. 
We chose this more complex distribution because we noticed that hidden state representations at different layers often contain distinct information. Specifically, the last layer is directly used for predicting the next token, while the second-to-last layer typically encodes the most abstract information about the current token. The zeroth layer represents the embedding layer, which contains context-unprocessed token information. Finally, we included half the depth of layers as a supplement.
We apply Grouped RMS Norm over the last dimension, assigning a group of scale parameters (initialized to $1$) to each slice along the second-to-last dimension. The normalized tensor is then reshaped into $I_{\mathrm{Cat}}\in \mathbb{R}^{bs\times\Abs{c}\times4d_h}$, which serves as the input to the Downsampler, a linear module with weights of $W_\mathrm{D}$ and no bias. The output $I_\mathrm{D}\in \mathbb{R}^{bs\times\Abs{c}\times d_h}$ is RMS-normalized again before being passed into the masked self-attention (MSA), which can be viewed as an additional $(L+1)$-th layer of the LLM. This design allows for easy integration with existing KV cache management frameworks. Formalized in Equation~\ref{eq:dec1}.
\begin{equation}
\label{eq:dec1}
\begin{aligned}
    I_{\mathrm{D}} = &
        (\mathrm{MSA} \cdot \mathrm{RMS} + \mathbb{I})\left(
            W_\mathrm{D}\cdot I_{\mathrm{Cat}}
        \right) \\
    I_{\mathrm{Cat}} = &
        \mathrm{GroupRMS} \left( \mathrm{HS}[0, L/2, L-1, L] \right)
\end{aligned}
\end{equation}
The Positional FFN is a linear projection from dimension $d_h$ to $l_d\cdot d_h$ with weights $W_\mathrm{P}$, effectively decomposing a representation into $l_d$ position-specific components with added biases $b_\mathrm{P}$. We argue that position-specific information in draft tokens should not be too simplistic, such as assigning a basic mask per position, yet using a full MLP for each position would be overly redundant. Therefore, we adopt this middle-ground approach. The number of position-related parameters is $l_d\cdot d_h^2$, which, while still quadratic in $d_h$, is smaller than methods like Medusa, which require at least $8\cdot l_d\cdot d_h^2$, placing our method on the more efficient end. The output of the Context Causal Attention stage is a tensor $D\in \mathbb{R}^{bs\times\Abs{c}\times l_d \times d_h}$. Formalized in Equation~\ref{eq:dec2}.
\begin{equation}
\label{eq:dec2}
    D = W_\mathrm{P} \cdot \mathrm{RMS} (I_{\mathrm{D}}) + b_\mathrm{P}
\end{equation}

\subsubsection{Draft Bi-directional Attention}

The Draft Bi-directional Attention layer applies self-attention mechanisms within the draft token sequence, utilizing a standard self-attention (SA) module with residual connections and a Swish Gated Linear Unit (SwiGLU) feedforward network. All components are normalized using RMS Normalization. The output $E\in \mathbb{R}^{bs\times\Abs{c}\times l_d \times d_h}$. Formalized in Equation~\ref{eq:enc}.
\begin{equation}
\label{eq:enc}
    E = (\mathrm{SwiGLU} \cdot \mathrm{RMS}+\mathbb{I})(
        (\mathrm{SA} \cdot \mathrm{RMS}+\mathbb{I})(D)
    )
\end{equation}
It is important to emphasize that the attention mechanism operates along the draft token dimension; that is, for a sequence of length $l_d$, the effective batch size becomes $bs \cdot \Abs{c}$. In our implementation, we observed that FlashAttention 2~\citep{DBLP:conf/iclr/Dao24, NEURIPS2022_67d57c32} cannot handle batch sizes larger than $4095$. To address this limitation, we partition the computation along the batch dimension, processing the attention in groups of $3072$ samples per batch segment.

\begin{table*}[ht]
    \centering
    \begin{tabular}{cc|cc|cc|cc|cc}
    \toprule
        \multirow{2}{*}{$bs$} & \multirow{2}{*}{$k$} 
            & \multicolumn{2}{c|}{W/o SD} 
            & \multicolumn{2}{c|}{HASS} 
            & \multicolumn{2}{c|}{EAGLE-3} 
            & \multicolumn{2}{c }{Ours} 
        \\
        &
            & $\kappa$ & TPS 
            & $\kappa$ & TPS 
            & $\kappa$ & TPS 
            & $\kappa$ & TPS 
        \\
    \midrule
        \multirow{3}{*}{1} 
        & 4 
            & 1    &   41 (1$\times$) 
            & 2.14 &   69 (1.70$\times$) 
            & 2.16 &   70 (1.73$\times$) 
            & 2.20 &   73 (1.78$\times$) 
        \\
        & 6 
            & 1    &   41 (1$\times$) 
            & 2.17 &   71 (1.74$\times$) 
            & 2.18 &   72 (1.75$\times$) 
            & 2.22 &   74 ({\bf 1.81}$\times$) 
        \\
        & 8 
            & 1    &   41 (1$\times$) 
            & 2.17 &   72 (1.75$\times$) 
            & 2.19 &   72 (1.76$\times$) 
            & 2.23 &   73 (1.80$\times$) 
        \\
    \midrule
        \multirow{3}{*}{4} 
        & 4  
            & 1    &  162 (1$\times$) 
            & 2.14 &  275 (1.70$\times$) 
            & 2.16 &  277 (1.71$\times$) 
            & 2.18 &  289 (1.78$\times$) 
        \\
        & 6  
            & 1    &  162 (1$\times$) 
            & 2.17 &  282 (1.74$\times$) 
            & 2.17 &  282 (1.73$\times$) 
            & 2.22 &  293 ({\bf 1.81}$\times$) 
        \\
        & 8  
            & 1    &  162 (1$\times$) 
            & 2.18 &  284 (1.75$\times$) 
            & 2.18 &  279 (1.72$\times$) 
            & 2.23 &  291 (1.80$\times$) 
        \\
    \midrule
        \multirow{3}{*}{16} 
        & 4  
            & 1    &  681 (1$\times$) 
            & 2.14 & 1164 (1.71$\times$) 
            & 2.16 & 1175 (1.72$\times$) 
            & 2.19 & 1212 (1.78$\times$) 
        \\
        & 6  
            & 1    &  681 (1$\times$)
            & 2.16 & 1190 (1.74$\times$) 
            & 2.17 & 1185 (1.74$\times$) 
            & 2.22 & 1233 ({\bf 1.81}$\times$) 
        \\
        & 8  
            & 1    &  681 (1$\times$)
            & 2.17 & 1189 (1.75$\times$) 
            & 2.17 & 1192 (1.75$\times$) 
            & 2.24 & 1220 (1.79$\times$) 
        \\
    \midrule
        \multirow{3}{*}{64} 
        & 4  
            & 1    & 2590 (1$\times$)
            & 2.13 & 4454 (1.72$\times$) 
            & 2.15 & 4429 (1.71$\times$) 
            & 2.19 & 4610 (1.78$\times$) 
        \\
        & 6  
            & 1    & 2590 (1$\times$)
            & 2.17 & 4530 (1.75$\times$) 
            & 2.17 & 4515 (1.74$\times$) 
            & 2.22 & 4688 ({\bf 1.81}$\times$) 
        \\
        & 8  
            & 1    & 2590 (1$\times$) 
            & 2.17 & 4541 (1.75$\times$) 
            & 2.18 & 4507 (1.74$\times$) 
            & 2.24 & 4610 (1.78$\times$) 
        \\
    \midrule
        \multirow{3}{*}{128} 
        & 4  
            & 1    & 5143 (1$\times$) 
            & 2.14 & 8800 (1.71$\times$) 
            & 2.16 & 8846 (1.72$\times$) 
            & 2.18 & 9154 (1.78$\times$) 
        \\
        & 6  
            & 1    & 5143 (1$\times$) 
            & 2.16 & 8956 (1.74$\times$) 
            & 2.17 & 8901 (1.73$\times$) 
            & 2.22 & 9308 ({\bf 1.81}$\times$) 
        \\
        & 8  
            & 1    & 5143 (1$\times$) 
            & 2.17 & 8945 (1.74$\times$) 
            & 2.16 & 8845 (1.72$\times$) 
            & 2.24 & 9206 (1.79$\times$) 
        \\
    \bottomrule
    \end{tabular}
    \caption{The comparison between SpecFormer and baselines under different batch size and settings. The baseline methods may underperform compared to their reported values, as we impose a constraint on the draft token budget.}
    \label{tab:tabmain}
\end{table*}

\subsection{Implementation Improvements}

\subsubsection{Efficient Grouped RMS Norm}

Through profiling, we found that RMS Normalization often becomes a performance bottleneck, primarily due to its significant consumption of CPU time slices. As a result, implementing Grouped RMS Norm with a loop-based approach tends to be inefficient. To address this, we customized a GPU kernel using Triton~\citep{DBLP:conf/pldi/TilletKC19} to implement the Grouped RMS Norm operation more efficiently.

\subsubsection{Intra-batch Gradient Accumulation}

We adopted the gradient accumulation strategy around the LM Head as proposed by \citeauthor{DBLP:conf/icml/GloeckleIRLS24}. Specifically, for each position $j\in \{1,2,\dots,l_d\}$, we compute the loss sequentially, rather than simultaneously. This is because the vocabulary size in modern language models often exceeds 128K, making the softmax very expensive in storage. 
Instead, we sequentially map each position’s hidden state to the vocabulary, compute gradients, and store them within the hidden states via backpropagation. 
Once gradients for all positions are computed, we continue the remaining backward pass together.

\section{Experiment}

\subsection{Priliminary Experiment}
We tested the GPU throughput achieved under different batch sizes and token counts to corroborate that the number of available tokens for speculative decoding is limited with large batch sizes. Our results are shown in Figure~\ref{fig:tpt}.

\subsection{Setups}
\subsubsection{Training Corpus}
We trained our model on the UltraChat-200K (UC, \citealp{ding2023enhancing}) dataset, which contains approximately 460K dialogue samples. Although the dataset itself is distilled from ChatGPT outputs, in our implementation, we opted to perform self-distillation~\citep{9381661, lasby2025sd2selfdistilledsparsedrafters} first. Specifically, we retained only the question parts from the original samples and regenerated the completions using the base LLM. This ensures that the distribution learned by the draft model strictly aligns with that of the base model. Our experiments demonstrate that this adjustment leads to performance improvements.

\subsubsection{Base LLM}
We selected foundation models from the Qwen and LLaMA families, including Qwen2.5-3B, Qwen3-8B, Qwen3-14B, and LLaMA-3.1-8B. Unlike many previous works, we did not adopt the Vicuna~\citep{DBLP:conf/nips/ZhengC00WZL0LXZ23} series. This decision is based on two considerations: First, both the Vicuna model and its training dataset (ShareGPT) are relatively outdated. Second, as a chat model built on early versions of LLaMA~\citep{DBLP:journals/corr/abs-2307-09288}, Vicuna uses a small vocabulary (about 32K). Vocabulary size is closely correlated with the difficulty of token prediction in draft generation—larger vocabularies increase prediction difficulty. Modern models typically use vocabularies exceeding 128K, with some, such as Gemma~\citep{DBLP:journals/corr/abs-2503-19786}, reaching 256K, making Vicuna unrepresentative of current LLMs.

\subsubsection{Evaluation}
Our evaluation set includes the test split of the UC dataset along with several popular benchmarks: MT-Bench~\citep{DBLP:conf/nips/ZhengC00WZL0LXZ23}, HumanEval~\citep{chen2021evaluating}, GSM8K~\citep{DBLP:journals/corr/abs-2110-14168}, Alpaca~\citep{alpaca}, and CNN/DM~\citep{DBLP:conf/acl/SeeLM17}. For reporting purposes, we present averaged results across this combined set, as there is no strong evidence suggesting performance varies significantly across these datasets in our no-regression setting. Since we focus on lossless LLM acceleration, correctness is not a concern—the model’s outputs remain identical before and after acceleration.

\subsubsection{Implementation}
Our method is implemented and trained using the {\it PyTorch} framework with few {\it Triton} and {\it FlashAttention} components. For inference, we leverage the {\it Medusa} decoding framework, as well as custom SD-compatible code based on the {\it HuggingFace Transformers}~\citep{DBLP:journals/corr/abs-1910-03771}. We conducted tests under various batch sizes, and report the theoretical speedup, efficiency factor $\kappa$, and actual speed gains. Detailed hyperparameters is given in Appendix~\ref{sec:appendix}.

\begin{table*}[ht]
    \centering
    \begin{tabular}{cc|ccc|ccc|ccc}
    \toprule
        \multirow{2}{*}{$bs$} & \multirow{2}{*}{$k$} 
         & \multicolumn{3}{c|}{W/o SD} & \multicolumn{3}{c|}{No-Self-Distill} & \multicolumn{3}{c}{Self-Distill}\\
        && $l_d$ & $\kappa$ & TPS & $l_d$ & $\kappa$ & TPS & $l_d$ & $\kappa$ & TPS \\
    \midrule
        1 & 8 & 1 & 1 & 32 (1.00$\times$) & 8 & 1.19 & 30 (0.94$\times$) & 8 & 1.90 & 56 (1.76$\times$) \\
    \bottomrule
    \end{tabular}
    \caption{The comparison between to use or not to use self-distillation.}
    \label{tab:sdono}
\end{table*}

\begin{table*}[ht]
    \centering
    \begin{tabular}{cc|ccc|ccc|ccc}
    \toprule
        \multirow{2}{*}{$bs$} & \multirow{2}{*}{$k$} 
            & \multicolumn{3}{c|}{Qwen3-4B} 
            & \multicolumn{3}{c|}{Qwen3-8B} 
            & \multicolumn{3}{c }{Qwen3-14B} 
        \\
        &
            & $\kappa$ & TPS & $\theta$
            & $\kappa$ & TPS & $\theta$
            & $\kappa$ & TPS & $\theta$
        \\
    \midrule
        \multirow{3}{*}{1} 
        & 0 
            & 1    & 30   (1.00$\times$) & 1
            & 1    & 31   (1.00$\times$) & 1
            & 1    & 26   (1.00$\times$) & 1
        \\
        & 4
            & 1.81 & 45   (1.50$\times$) & 1.21
            & 1.74 & 45   (1.45$\times$) & 1.20
            & 1.71 & 38   (1.46$\times$) & 1.17
        \\
        & 8 
            & 1.81 & 46   (1.54$\times$) & 1.18
            & 1.76 & 46   (1.49$\times$) & 1.18
            & 1.72 & 39   (1.46$\times$) & 1.18
        \\
    \midrule
        \multirow{3}{*}{4} 
        & 0  
            & 1    & 147  (1.00$\times$) & 1
            & 1    & 120  (1.00$\times$) & 1
            & 1    & 105  (1.00$\times$) & 1
        \\
        & 4  
            & 1.84 & 224  (1.53$\times$) & 1.20
            & 1.76 & 178  (1.48$\times$) & 1.19
            & 1.71 & 157  (1.49$\times$) & 1.14
        \\
        & 8 
            & 1.86 & 227  (1.56$\times$) & 1.19
            & 1.76 & 182  (1.49$\times$) & 1.18
            & 1.72 & 154  (1.47$\times$) & 1.17
        \\
    \midrule
        \multirow{3}{*}{16} 
        & 0  
            & 1    & 588  (1.00$\times$) & 1
            & 1    & 488  (1.00$\times$) & 1
            & 1    & 436  (1.00$\times$) & 1
        \\
        & 4  
            & 1.84 & 899  (1.53$\times$) & 1.20
            & 1.76 & 726  (1.49$\times$) & 1.18
            & 1.71 & 636  (1.47$\times$) & 1.16
        \\
        & 8  
            & 1.86 & 917  (1.56$\times$) & 1.19
            & 1.77 & 726  (1.49$\times$) & 1.19
            & 1.72 & 639  (1.46$\times$) & 1.18
        \\
    \midrule
        \multirow{3}{*}{64} 
        & 0  
            & 1    & 2346 (1.00$\times$) & 1
            & 1    & 1904 (1.00$\times$) & 1
            & 1    & 1713 (1.00$\times$) & 1
        \\
        & 2  
            & 1.72 & 3435 (1.46$\times$) & 1.18
            & 1.68 & 2734 (1.44$\times$) & 1.17
            & 1.64 & 2454 (1.41$\times$) & 1.16
        \\
        & 4  
            & 1.84 & 3621 (1.53$\times$) & 1.20
            & 1.75 & 2834 (1.48$\times$) & 1.18
            & 1.71 & 2524 (1.47$\times$) & 1.16
        \\
    \midrule
        \multirow{3}{*}{128} 
        & 0  
            & 1    & 4582 (1.00$\times$) & 1
            & 1    & 3882 (1.00$\times$) & 1
            & 1    & 3458 (1.00$\times$) & 1
        \\
        & 2  
            & 1.73 & 6725 (1.47$\times$) & 1.18
            & 1.68 & 5586 (1.43$\times$) & 1.17
            & 1.64 & 4834 (1.41$\times$) & 1.16
        \\
        & 4  
            & 1.84 & 7263 (1.53$\times$) & 1.20
            & 1.75 & 5761 (1.48$\times$) & 1.18
            & 1.71 & 5090 (1.47$\times$) & 1.16
        \\
    \bottomrule
    \end{tabular}
    \caption{The comparison between our proposed method SpecFormer and baselines under size of base LLMs.}
    \label{tab:modelsz}
\end{table*}

\subsection{Throughput Comparison}

We constrain the available draft token budget to a relatively small value and then evaluate the system's throughput under varying batch sizes.
We measure the throughput of our method using tokens per second (TPS), as shown in Table~\ref{tab:tabmain}. We observe that our approach consistently outperforms the baseline methods. Notably, the baselines do not reach their reported performance levels in our setting because we constrain the available token budget to simulate scenarios with limited computational redundancy, such as those arising in large-batch inference. In contrast, our method achieves high throughput without relying on a large number of draft tokens, owing to its superior predictive capability. Furthermore, we evaluate the conversion rate from $\kappa$-to-TPS, and find that our method exhibits a higher conversion efficiency. This is primarily because our design adopts a non-autoregressive formulation, which results in higher arithmetic intensity and lower average per-token overhead, thereby improving overall efficiency.

\subsection{Special Case Study}

\subsubsection{Self Distillation}

We evaluate the impact of self-distillation by comparing models trained with and without it on Qwen2.5-3B. Specifically, we first train an {\it No-Self-Distill} model using the original UC-200K dialogue dataset. Then, we apply self-distillation by retaining only the prompt side of each dialogue and generating completions using the base LLM, which are subsequently used to train the {\it Self-Distill} model. Notably, the self-distilled dataset is smaller in size, as it contains fewer dialogue turns.

The $\kappa$ value and acceleration performance are reported in Table~\ref{tab:sdono}. We observe that without self-distillation, the model demonstrates negligible acceleration, as the learned token distribution does not originate from the base model, but rather from a different teacher model. While traditional distillation may partially mitigate this issue, we argue that self-distillation remains a necessary step, particularly in light of modern deployment frameworks like {\it vLLM}, which offer highly efficient offline inference and make strict alignment with the base model's output even more critical.

\subsubsection{Base LLM Size}

To investigate the performance gains of our architecture under speculative decoding across different model sizes, we conducted experiments on the Qwen-3 series, including 4B, 8B, and 14B variants—covering a representative range of commonly used model scales. The acceleration results across these models are presented in Table~\ref{tab:modelsz}. We also calculate the $\kappa$-to-TPS conversion ratio $\theta$ to measure how the draft module itself impact the efficiency.

We observe that as the model size increases, the predictor’s ability to accurately guess future tokens are weakened, resulting in less acceleration gains. For instance, the 4B model achieves a speedup of 1.56×, whereas the 14B model sees a reduced speedup of 1.47×. However, we also find that larger models exhibit a more favorable $\theta$, meaning that the relative overhead introduced by the predictor is smaller. This can be attributed to two main reasons: The increased number of layers in larger models leads to a smaller parameter percentage for the predictor, and the larger weight matrices in big models dilute the overhead from scheduling. Overall, these results demonstrate that our method remains applicable across various model sizes, although it shows particularly strong benefits on smaller models.

\subsection{Module Ablation Study}

We conducted ablation studies on the Qwen3-4B model with $k=8$. Our analysis focused on whether bidirectional attention improves the model’s capability, whether applying a naïve linear transformation for positional encoding enhances performance, and the model’s capability under a wider architecture. We modify each module individually and ran experiments accordingly, and the results are shown in Table~\ref{tab:abla}.

\begin{table}[ht]
    \centering
    \begin{tabular}{c|c|ccc}
        \toprule
        Method & SpecFormer & -Pos & +Att Mask & +Larger \\
        \midrule
        $\kappa$ & 1.81 & 1.77 & 1.80 & 1.91 \\
        \bottomrule
    \end{tabular}
    \caption{Ablation Study}
    \label{tab:abla}
\end{table}

We note the following: 1. Bidirectional attention improves the model but not substantial. However, considering that its impact on inference time is negligible, we chose to retain this structure. 2. The positional FFN contributes a large portion of improvement, which is expected as it accounts for a considerable amount of parameters. 3. Larger model size leads to significant performance gains. This suggests that scaling up the model can offset the negative impact of using a deeper base model, whose total parameter count is typically larger, on the proportion of parameters allocated to the draft model.

\section{Conclusion}
% We first analyze that the batch execution environment imposes constraints on the effectiveness of speculative decoding by decreasing the idle computational resources. 
In this work, we first identified the dilemma of SD under modern inference services: batch processing compresses available extra computational resources, thereby limiting the draft size.
% and making broader exploration difficult 
Furthermore, the large number of position-dependent parameters in current draft model architectures hinders their ability to scale effectively, making it challenging to increase parameter size to improve prediction accuracy.
Then we proposed a novel SD method for LLMs, termed SpecFormer, which leverages two types of attention mechanisms operating along different dimensions, one unidirectional and one bidirectional. This design enables efficient parallel generation of future tokens while extracting information from the full context, resulting in a more capable draft model. Consequently, our approach maintains high prediction accuracy under a limited draft token budget.
We further conduct experiments across varying batch sizes, demonstrating that our method sustains comparable performance as batch size increases. Lastly, evaluations on models of different scales confirm the general applicability of our approach across a broad range of LLM configurations.

\section{Acknowledgments}

This work was supported by the National Natural Science Foundation of
China (No. 62306216), the Fundamental Research Funds for the Central
Universities (No.2042025kf0026), the Technology Innovation Program of Hubei Province (Grant No. 2024BAB043), and the Xiaomi Open-Competition Research Program.


% \section*{Limitations}

% Our method has several limitations that highlight possible directions for future work. First, it requires training, even though we only train the draft-module, which imposes relatively modest demands in terms of compute and supervision, the inclusion of a self-distillation stage still entails a nontrivial number of GPU-hours. Fully training-free approaches may represent a promising avenue for further research.

% Moreover, our method, as with any non-autoregressive decoding strategy, faces inherent challenges when integrated with prefix tree structures, where autoregressive methods currently hold a clear advantage. Developing more effective and efficient mechanisms to couple non-autoregressive predictors with prefix-based verification remains an open and valuable research problem.

% \section*{Ethics Statement}

% This work does not involve the collection or use of any personally identifiable data, human subjects, or sensitive information. All experiments are conducted using publicly available datasets and open-source models. We adhere to the principles of responsible AI research, including transparency, reproducibility, and fairness. Any use of large language models complies with the respective licensing terms. Our proposed methods are intended for research purposes only and should be deployed with care to avoid misuse or unintended consequences.

\bibliography{aaai2026}

@article{DBLP:journals/corr/abs-2303-08774,
  author       = {OpenAI},
  title        = {{GPT-4} Technical Report},
  journal      = {CoRR},
  volume       = {abs/2303.08774},
  year         = {2023},
  url          = {https://doi.org/10.48550/arXiv.2303.08774},
  doi          = {10.48550/ARXIV.2303.08774},
  eprinttype    = {arXiv},
  eprint       = {2303.08774},
  bibsource    = {dblp computer science bibliography, https://dblp.org}
}

@article{DBLP:journals/corr/abs-2407-18003,
  author       = {Luohe Shi and
                  Hongyi Zhang and
                  Yao Yao and
                  Zuchao Li and
                  Hai Zhao},
  title        = {Keep the Cost Down: {A} Review on Methods to Optimize LLM' s
                  KV-Cache Consumption},
  journal      = {CoRR},
  volume       = {abs/2407.18003},
  year         = {2024},
  url          = {https://doi.org/10.48550/arXiv.2407.18003},
  doi          = {10.48550/ARXIV.2407.18003},
  eprinttype    = {arXiv},
  eprint       = {2407.18003},
  bibsource    = {dblp computer science bibliography, https://dblp.org}
}

@article{DBLP:journals/cacm/WilliamsWP09,
  author       = {Samuel Williams and
                  Andrew Waterman and
                  David A. Patterson},
  title        = {Roofline: an insightful visual performance model for multicore architectures},
  journal      = {Commun. {ACM}},
  volume       = {52},
  number       = {4},
  pages        = {65--76},
  year         = {2009},
  url          = {https://doi.org/10.1145/1498765.1498785},
  doi          = {10.1145/1498765.1498785},
  bibsource    = {dblp computer science bibliography, https://dblp.org}
}

@inproceedings {298687,
author = {Yinmin Zhong and Shengyu Liu and Junda Chen and Jianbo Hu and Yibo Zhu and et al.},
title = {{DistServe}: Disaggregating Prefill and Decoding for Goodput-optimized Large Language Model Serving},
booktitle = {18th USENIX Symposium on Operating Systems Design and Implementation (OSDI 24)},
year = {2024},
isbn = {978-1-939133-40-3},
address = {Santa Clara, CA},
pages = {193--210},
url = {https://www.usenix.org/conference/osdi24/presentation/zhong-yinmin},
publisher = {USENIX Association},
month = jul
}

@inproceedings {280922,
author = {Gyeong-In Yu and Joo Seong Jeong and Geon-Woo Kim and Soojeong Kim and Byung-Gon Chun},
title = {Orca: A Distributed Serving System for {Transformer-Based} Generative Models},
booktitle = {16th USENIX Symposium on Operating Systems Design and Implementation (OSDI 22)},
year = {2022},
isbn = {978-1-939133-28-1},
address = {Carlsbad, CA},
pages = {521--538},
url = {https://www.usenix.org/conference/osdi22/presentation/yu},
publisher = {USENIX Association},
month = jul
}

@inproceedings{10.1145/3600006.3613165,
author = {Kwon, Woosuk and Li, Zhuohan and Zhuang, Siyuan and Sheng, Ying and Zheng, Lianmin and et al.},
title = {Efficient Memory Management for Large Language Model Serving with PagedAttention},
year = {2023},
isbn = {9798400702297},
publisher = {Association for Computing Machinery},
address = {New York, NY, USA},
url = {https://doi.org/10.1145/3600006.3613165},
doi = {10.1145/3600006.3613165},
booktitle = {Proceedings of the 29th Symposium on Operating Systems Principles},
pages = {611–626},
numpages = {16},
location = {Koblenz, Germany},
series = {SOSP '23}
}

@article{DBLP:journals/corr/abs-2410-14257,
  author       = {Zhibin Wang and
                  Shipeng Li and
                  Yuhang Zhou and
                  Xue Li and
                  Rong Gu and
                  et al.},
  title        = {Revisiting {SLO} and Goodput Metrics in {LLM} Serving},
  journal      = {CoRR},
  volume       = {abs/2410.14257},
  year         = {2024},
  url          = {https://doi.org/10.48550/arXiv.2410.14257},
  doi          = {10.48550/ARXIV.2410.14257},
  eprinttype    = {arXiv},
  eprint       = {2410.14257},
  bibsource    = {dblp computer science bibliography, https://dblp.org}
}

@inproceedings{xia-etal-2024-unlocking,
    title = "Unlocking Efficiency in Large Language Model Inference: A Comprehensive Survey of Speculative Decoding",
    author = "Xia, Heming and Yang, Zhe and Dong, Qingxiu and Wang, Peiyi and Li, Yongqi  and et al.",
    editor = "Ku, Lun-Wei and Martins, Andre and Srikumar, Vivek",
    booktitle = "Findings of the Association for Computational Linguistics ACL 2024",
    month = aug,
    year = "2024",
    address = "Bangkok, Thailand and virtual meeting",
    publisher = "Association for Computational Linguistics",
    url = "https://aclanthology.org/2024.findings-acl.456",
    doi = "10.18653/v1/2024.findings-acl.456",
    pages = "7655--7671",
}

@inproceedings{NEURIPS2018_c4127b91,
 author = {Stern, Mitchell and Shazeer, Noam and Uszkoreit, Jakob},
 booktitle = {Advances in Neural Information Processing Systems},
 editor = {S. Bengio and H. Wallach and H. Larochelle and K. Grauman and N. Cesa-Bianchi and R. Garnett},
 pages = {},
 publisher = {Curran Associates, Inc.},
 title = {Blockwise Parallel Decoding for Deep Autoregressive Models},
 url = {https://proceedings.neurips.cc/paper_files/paper/2018/file/c4127b9194fe8562c64dc0f5bf2c93bc-Paper.pdf},
 volume = {31},
 year = {2018}
}

@article{DBLP:journals/corr/abs-2502-19732,
  author       = {Yunhai Hu and
                  Zining Liu and
                  Zhenyuan Dong and
                  Tianfan Peng and
                  Bradley McDanel and
                  Sai Qian Zhang},
  title        = {Speculative Decoding and Beyond: An In-Depth Survey of Techniques},
  journal      = {CoRR},
  volume       = {abs/2502.19732},
  year         = {2025},
  url          = {https://doi.org/10.48550/arXiv.2502.19732},
  doi          = {10.48550/ARXIV.2502.19732},
  eprinttype    = {arXiv},
  eprint       = {2502.19732},
  bibsource    = {dblp computer science bibliography, https://dblp.org}
}

@article{qwen2.5,
    title   = {Qwen2.5 Technical Report}, 
    author  = {An Yang and Baosong Yang and Beichen Zhang and Binyuan Hui and Bo Zheng and et al.},
    journal = {arXiv preprint arXiv:2412.15115},
    year    = {2024}
}

@inproceedings{NEURIPS2023_7b97adea,
 author = {Kim, Sehoon and Mangalam, Karttikeya and Moon, Suhong and Malik, Jitendra and Mahoney, Michael W and et al.},
 booktitle = {Advances in Neural Information Processing Systems},
 editor = {A. Oh and T. Naumann and A. Globerson and K. Saenko and M. Hardt and S. Levine},
 pages = {39236--39256},
 publisher = {Curran Associates, Inc.},
 title = {Speculative Decoding with Big Little Decoder},
 url = {https://proceedings.neurips.cc/paper_files/paper/2023/file/7b97adeafa1c51cf65263459ca9d0d7c-Paper-Conference.pdf},
 volume = {36},
 year = {2023}
}

@inproceedings{DBLP:conf/emnlp/Xia0WCWS23,
  author       = {Heming Xia and
                  Tao Ge and
                  Peiyi Wang and
                  Si{-}Qing Chen and
                  Furu Wei and
                  Zhifang Sui},
  editor       = {Houda Bouamor and
                  Juan Pino and
                  Kalika Bali},
  title        = {Speculative Decoding: Exploiting Speculative Execution for Accelerating
                  Seq2seq Generation},
  booktitle    = {Findings of the Association for Computational Linguistics: {EMNLP}
                  2023, Singapore, December 6-10, 2023},
  pages        = {3909--3925},
  publisher    = {Association for Computational Linguistics},
  year         = {2023},
  url          = {https://doi.org/10.18653/v1/2023.findings-emnlp.257},
  doi          = {10.18653/V1/2023.FINDINGS-EMNLP.257},
  bibsource    = {dblp computer science bibliography, https://dblp.org}
}

@article{DBLP:journals/corr/abs-2405-19715,
  author       = {Kaixuan Huang and
                  Xudong Guo and
                  Mengdi Wang},
  title        = {SpecDec++: Boosting Speculative Decoding via Adaptive Candidate Lengths},
  journal      = {CoRR},
  volume       = {abs/2405.19715},
  year         = {2024},
  url          = {https://doi.org/10.48550/arXiv.2405.19715},
  doi          = {10.48550/ARXIV.2405.19715},
  eprinttype    = {arXiv},
  eprint       = {2405.19715},
  bibsource    = {dblp computer science bibliography, https://dblp.org}
}

@inproceedings{DBLP:conf/iclr/ZhouLRMRKKA24,
  author       = {Yongchao Zhou and
                  Kaifeng Lyu and
                  Ankit Singh Rawat and
                  Aditya Krishna Menon and
                  Afshin Rostamizadeh and
                  Sanjiv Kumar and
                  et al.},
  title        = {DistillSpec: Improving Speculative Decoding via Knowledge Distillation},
  booktitle    = {The Twelfth International Conference on Learning Representations,
                  {ICLR} 2024, Vienna, Austria, May 7-11, 2024},
  publisher    = {OpenReview.net},
  year         = {2024},
  url          = {https://openreview.net/forum?id=rsY6J3ZaTF},
  bibsource    = {dblp computer science bibliography, https://dblp.org}
}

@inproceedings{
bachmann2025judge,
title={Judge Decoding: Faster Speculative Sampling Requires Going Beyond Model Alignment},
author={Gregor Bachmann and Sotiris Anagnostidis and Albert Pumarola and Markos Georgopoulos and Artsiom Sanakoyeu and et al.},
booktitle={The Thirteenth International Conference on Learning Representations},
year={2025},
url={https://openreview.net/forum?id=mtSSFiqW6y}
}

@misc{li2025eaglespeculativesamplingrequires,
      title={EAGLE: Speculative Sampling Requires Rethinking Feature Uncertainty}, 
      author={Yuhui Li and Fangyun Wei and Chao Zhang and Hongyang Zhang},
      year={2025},
      eprint={2401.15077},
      archivePrefix={arXiv},
      primaryClass={cs.LG},
      url={https://arxiv.org/abs/2401.15077}, 
}

@article{Gao_Xie_Xiang_Ji_2025, title={Falcon: Faster and Parallel Inference of Large Language Models Through Enhanced Semi-Autoregressive Drafting and Custom-Designed Decoding Tree}, volume={39}, url={https://ojs.aaai.org/index.php/AAAI/article/view/34566}, DOI={10.1609/aaai.v39i22.34566}, abstractNote={Striking an optimal balance between minimal drafting latency and high speculation accuracy to enhance the inference speed of Large Language Models remains a significant challenge in speculative decoding. In this paper, we introduce Falcon, an innovative semi-autoregressive speculative decoding framework fashioned to augment both the drafter’s parallelism and output quality. Falcon incorporates the Coupled Sequential Glancing Distillation technique, which fortifies inter-token dependencies within the same block, leading to increased speculation accuracy. We offer a comprehensive theoretical analysis to illuminate the underlying mechanisms. Additionally, we introduce a Custom-Designed Decoding Tree, which permits the drafter to generate multiple tokens in a single forward pass and accommodates multiple forward passes as needed, thereby boosting the number of drafted tokens and significantly improving the overall acceptance rate. Comprehensive evaluations on benchmark datasets such as MT-Bench, HumanEval, and GSM8K demonstrate Falcon’s superior acceleration capabilities. The framework achieves a lossless speedup ratio ranging from 2.91x to 3.51x when tested on the Vicuna and LLaMA2-Chat model series. These results outstrip existing speculative decoding methods for LLMs, including Eagle, Medusa, Lookahead, SPS, and PLD, while maintaining a compact drafter architecture equivalent to merely two Transformer layers.}, number={22}, journal={Proceedings of the AAAI Conference on Artificial Intelligence}, author={Gao, Xiangxiang and Xie, Weisheng and Xiang, Yiwei and Ji, Feng}, year={2025}, month={Apr.}, pages={23933-23941} }

@inproceedings{
zhang2025learning,
title={Learning Harmonized Representations for Speculative Sampling},
author={Lefan Zhang and Xiaodan Wang and Yanhua Huang and Ruiwen Xu},
booktitle={The Thirteenth International Conference on Learning Representations},
year={2025},
url={https://openreview.net/forum?id=T9u56s7mbk}
}

@article{DBLP:journals/corr/abs-2412-19437,
  author       = {DeepSeek{-}AI},
  title        = {DeepSeek-V3 Technical Report},
  journal      = {CoRR},
  volume       = {abs/2412.19437},
  year         = {2024},
  url          = {https://doi.org/10.48550/arXiv.2412.19437},
  doi          = {10.48550/ARXIV.2412.19437},
  eprinttype    = {arXiv},
  eprint       = {2412.19437},
  bibsource    = {dblp computer science bibliography, https://dblp.org}
}

@inproceedings{DBLP:conf/icml/GloeckleIRLS24,
  author       = {Fabian Gloeckle and
                  Badr Youbi Idrissi and
                  Baptiste Rozi{\`{e}}re and
                  David Lopez{-}Paz and
                  Gabriel Synnaeve},
  title        = {Better {\&} Faster Large Language Models via Multi-token Prediction},
  booktitle    = {Forty-first International Conference on Machine Learning, {ICML} 2024,
                  Vienna, Austria, July 21-27, 2024},
  publisher    = {OpenReview.net},
  year         = {2024},
  url          = {https://openreview.net/forum?id=pEWAcejiU2},
  bibsource    = {dblp computer science bibliography, https://dblp.org}
}

@inproceedings{DBLP:conf/icml/CaiLGPLCD24,
  author       = {Tianle Cai and
                  Yuhong Li and
                  Zhengyang Geng and
                  Hongwu Peng and
                  Jason D. Lee and
                  et al.},
  title        = {Medusa: Simple {LLM} Inference Acceleration Framework with Multiple
                  Decoding Heads},
  booktitle    = {Forty-first International Conference on Machine Learning, {ICML} 2024,
                  Vienna, Austria, July 21-27, 2024},
  publisher    = {OpenReview.net},
  year         = {2024},
  url          = {https://openreview.net/forum?id=PEpbUobfJv},
  bibsource    = {dblp computer science bibliography, https://dblp.org}
}

@article{DBLP:journals/corr/abs-2503-01840,
  author       = {Yuhui Li and
                  Fangyun Wei and
                  Chao Zhang and
                  Hongyang Zhang},
  title        = {{EAGLE-3:} Scaling up Inference Acceleration of Large Language Models
                  via Training-Time Test},
  journal      = {CoRR},
  volume       = {abs/2503.01840},
  year         = {2025},
  url          = {https://doi.org/10.48550/arXiv.2503.01840},
  doi          = {10.48550/ARXIV.2503.01840},
  eprinttype    = {arXiv},
  eprint       = {2503.01840},
  bibsource    = {dblp computer science bibliography, https://dblp.org}
}

@inproceedings{DBLP:conf/pldi/TilletKC19,
  author       = {Philippe Tillet and
                  Hsiang{-}Tsung Kung and
                  David D. Cox},
  editor       = {Tim Mattson and
                  Abdullah Muzahid and
                  Armando Solar{-}Lezama},
  title        = {Triton: an intermediate language and compiler for tiled neural network
                  computations},
  booktitle    = {Proceedings of the 3rd {ACM} {SIGPLAN} International Workshop on Machine
                  Learning and Programming Languages, MAPL@PLDI 2019, Phoenix, AZ, USA,
                  June 22, 2019},
  pages        = {10--19},
  publisher    = {{ACM}},
  year         = {2019},
  url          = {https://doi.org/10.1145/3315508.3329973},
  doi          = {10.1145/3315508.3329973},
  bibsource    = {dblp computer science bibliography, https://dblp.org}
}

@inproceedings{DBLP:conf/iclr/Dao24,
  author       = {Tri Dao},
  title        = {FlashAttention-2: Faster Attention with Better Parallelism and Work
                  Partitioning},
  booktitle    = {The Twelfth International Conference on Learning Representations,
                  {ICLR} 2024, Vienna, Austria, May 7-11, 2024},
  publisher    = {OpenReview.net},
  year         = {2024},
  url          = {https://openreview.net/forum?id=mZn2Xyh9Ec},
  bibsource    = {dblp computer science bibliography, https://dblp.org}
}

@inproceedings{NEURIPS2022_67d57c32,
 author = {Dao, Tri and Fu, Dan and Ermon, Stefano and Rudra, Atri and R\'{e}, Christopher},
 booktitle = {Advances in Neural Information Processing Systems},
 editor = {S. Koyejo and S. Mohamed and A. Agarwal and D. Belgrave and K. Cho and A. Oh},
 pages = {16344--16359},
 publisher = {Curran Associates, Inc.},
 title = {FlashAttention: Fast and Memory-Efficient Exact Attention with IO-Awareness},
 url = {https://proceedings.neurips.cc/paper_files/paper/2022/file/67d57c32e20fd0a7a302cb81d36e40d5-Paper-Conference.pdf},
 volume = {35},
 year = {2022}
}

@ARTICLE{9381661,

  author={Zhang, Linfeng and Bao, Chenglong and Ma, Kaisheng},

  journal={IEEE Transactions on Pattern Analysis and Machine Intelligence}, 

  title={Self-Distillation: Towards Efficient and Compact Neural Networks}, 

  year={2022},

  volume={44},

  number={8},

  pages={4388-4403},

  doi={10.1109/TPAMI.2021.3067100}}

@misc{ding2023enhancing,
      title={Enhancing Chat Language Models by Scaling High-quality Instructional Conversations}, 
      author={Ning Ding and Yulin Chen and Bokai Xu and Yujia Qin and Zhi Zheng and et al.},
      year={2023},
      eprint={2305.14233},
      archivePrefix={arXiv},
      primaryClass={cs.CL}
}

@misc{lasby2025sd2selfdistilledsparsedrafters,
      title={SD$^2$: Self-Distilled Sparse Drafters}, 
      author={Mike Lasby and Nish Sinnadurai and Valavan Manohararajah and Sean Lie and Vithursan Thangarasa},
      year={2025},
      eprint={2504.08838},
      archivePrefix={arXiv},
      primaryClass={cs.CL},
      url={https://arxiv.org/abs/2504.08838}, 
}

@inproceedings{DBLP:conf/nips/ZhengC00WZL0LXZ23,
  author       = {Lianmin Zheng and
                  Wei{-}Lin Chiang and
                  Ying Sheng and
                  Siyuan Zhuang and
                  et al.},
  editor       = {Alice Oh and
                  Tristan Naumann and
                  Amir Globerson and
                  Kate Saenko and
                  Moritz Hardt and
                  Sergey Levine},
  title        = {Judging LLM-as-a-Judge with MT-Bench and Chatbot Arena},
  booktitle    = {Advances in Neural Information Processing Systems 36: Annual Conference
                  on Neural Information Processing Systems 2023, NeurIPS 2023, New Orleans,
                  LA, USA, December 10 - 16, 2023},
  year         = {2023},
  url          = {http://papers.nips.cc/paper\_files/paper/2023/hash/91f18a1287b398d378ef22505bf41832-Abstract-Datasets\_and\_Benchmarks.html},
  bibsource    = {dblp computer science bibliography, https://dblp.org}
}

@article{DBLP:journals/corr/abs-2307-09288,
  author       = {Hugo Touvron and
                  Louis Martin and
                  Kevin Stone and
                  Peter Albert and
                  Amjad Almahairi and
                  et al.},
  title        = {Llama 2: Open Foundation and Fine-Tuned Chat Models},
  journal      = {CoRR},
  volume       = {abs/2307.09288},
  year         = {2023},
  url          = {https://doi.org/10.48550/arXiv.2307.09288},
  doi          = {10.48550/ARXIV.2307.09288},
  eprinttype    = {arXiv},
  eprint       = {2307.09288},
  bibsource    = {dblp computer science bibliography, https://dblp.org}
}

@article{DBLP:journals/corr/abs-2503-19786,
  author       = {Aishwarya Kamath and
                  Johan Ferret and
                  Shreya Pathak and
                  Nino Vieillard and
                  Ramona Merhej and
                  et al.},
  title        = {Gemma 3 Technical Report},
  journal      = {CoRR},
  volume       = {abs/2503.19786},
  year         = {2025},
  url          = {https://doi.org/10.48550/arXiv.2503.19786},
  doi          = {10.48550/ARXIV.2503.19786},
  eprinttype    = {arXiv},
  eprint       = {2503.19786},
  bibsource    = {dblp computer science bibliography, https://dblp.org}
}

@article{chen2021evaluating,
  author       = {Mark Chen and
                  Jerry Tworek and
                  Heewoo Jun and
                  Qiming Yuan and
                  Henrique Pond{\'{e}} de Oliveira Pinto and
                  et al.},
  title        = {Evaluating Large Language Models Trained on Code},
  journal      = {CoRR},
  volume       = {abs/2107.03374},
  year         = {2021},
  url          = {https://arxiv.org/abs/2107.03374},
  eprinttype    = {arXiv},
  eprint       = {2107.03374},
  bibsource    = {dblp computer science bibliography, https://dblp.org}
}

@article{DBLP:journals/corr/abs-2110-14168,
  author       = {Karl Cobbe and
                  Vineet Kosaraju and
                  Mohammad Bavarian and
                  Mark Chen and
                  Heewoo Jun and
                  et al.},
  title        = {Training Verifiers to Solve Math Word Problems},
  journal      = {CoRR},
  volume       = {abs/2110.14168},
  year         = {2021},
  url          = {https://arxiv.org/abs/2110.14168},
  eprinttype    = {arXiv},
  eprint       = {2110.14168},
  bibsource    = {dblp computer science bibliography, https://dblp.org}
}

@misc{alpaca,
  author = {Rohan Taori and Ishaan Gulrajani and Tianyi Zhang and Yann Dubois and Xuechen Li and Carlos Guestrin and Percy Liang and Tatsunori B. Hashimoto },
  title = {Stanford Alpaca: An Instruction-following LLaMA model},
  year = {2023},
  publisher = {GitHub},
  journal = {GitHub repository},
  howpublished = {\url{https://github.com/tatsu-lab/stanford_alpaca}},
}

@inproceedings{DBLP:conf/acl/SeeLM17,
  author       = {Abigail See and
                  Peter J. Liu and
                  Christopher D. Manning},
  editor       = {Regina Barzilay and
                  Min{-}Yen Kan},
  title        = {Get To The Point: Summarization with Pointer-Generator Networks},
  booktitle    = {Proceedings of the 55th Annual Meeting of the Association for Computational
                  Linguistics, {ACL} 2017, Vancouver, Canada, July 30 - August 4, Volume
                  1: Long Papers},
  pages        = {1073--1083},
  publisher    = {Association for Computational Linguistics},
  year         = {2017},
  url          = {https://doi.org/10.18653/v1/P17-1099},
  doi          = {10.18653/V1/P17-1099},
  bibsource    = {dblp computer science bibliography, https://dblp.org}
}

@article{DBLP:journals/corr/abs-1910-03771,
  author       = {Thomas Wolf and
                  Lysandre Debut and
                  Victor Sanh and
                  Julien Chaumond and
                  Clement Delangue and
                  et al.},
  title        = {HuggingFace's Transformers: State-of-the-art Natural Language Processing},
  journal      = {CoRR},
  volume       = {abs/1910.03771},
  year         = {2019},
  url          = {http://arxiv.org/abs/1910.03771},
  eprinttype    = {arXiv},
  eprint       = {1910.03771},
  bibsource    = {dblp computer science bibliography, https://dblp.org}
}

@inproceedings{chen-etal-2025-faster,
    title = "Faster In-Context Learning for {LLM}s via N-Gram Trie Speculative Decoding",
    author = "Chen, Jinglin  and
      Li, Qiwei  and
      Li, Zuchao  and
      Qi, Baoyuan  and
      Guoming, Liu  and
      Ai, Haojun  and
      Zhao, Hai  and
      Wang, Ping",
    editor = "Christodoulopoulos, Christos  and
      Chakraborty, Tanmoy  and
      Rose, Carolyn  and
      Peng, Violet",
    booktitle = "Proceedings of the 2025 Conference on Empirical Methods in Natural Language Processing",
    month = nov,
    year = "2025",
    address = "Suzhou, China",
    publisher = "Association for Computational Linguistics",
    url = "https://aclanthology.org/2025.emnlp-main.911/",
    doi = "10.18653/v1/2025.emnlp-main.911",
    pages = "18051--18062",
    ISBN = "979-8-89176-332-6"
}

@inproceedings{guo-etal-2025-tom,
    title = "{T}o{M}: Leveraging Tree-oriented {M}ap{R}educe for Long-Context Reasoning in Large Language Models",
    author = "Guo, Jiani  and
      Li, Zuchao  and
      Wu, Jie  and
      Wang, Qianren  and
      Li, Yun  and
      Zhang, Lefei  and
      Zhao, Hai  and
      Yang, Yujiu",
    editor = "Christodoulopoulos, Christos  and
      Chakraborty, Tanmoy  and
      Rose, Carolyn  and
      Peng, Violet",
    booktitle = "Proceedings of the 2025 Conference on Empirical Methods in Natural Language Processing",
    month = nov,
    year = "2025",
    address = "Suzhou, China",
    publisher = "Association for Computational Linguistics",
    url = "https://aclanthology.org/2025.emnlp-main.899/",
    doi = "10.18653/v1/2025.emnlp-main.899",
    pages = "17804--17823",
    ISBN = "979-8-89176-332-6"
}

@inproceedings{yang-etal-2025-xquant,
    title = "{XQ}uant: Achieving Ultra-Low Bit {KV} Cache Quantization with Cross-Layer Compression",
    author = "Yang, Haoqi  and
      Yao, Yao  and
      Li, Zuchao  and
      Qi, Baoyuan  and
      Guoming, Liu  and
      Zhao, Hai",
    editor = "Christodoulopoulos, Christos  and
      Chakraborty, Tanmoy  and
      Rose, Carolyn  and
      Peng, Violet",
    booktitle = "Proceedings of the 2025 Conference on Empirical Methods in Natural Language Processing",
    month = nov,
    year = "2025",
    address = "Suzhou, China",
    publisher = "Association for Computational Linguistics",
    url = "https://aclanthology.org/2025.emnlp-main.494/",
    doi = "10.18653/v1/2025.emnlp-main.494",
    pages = "9796--9811",
    ISBN = "979-8-89176-332-6"
}

@article{DBLP:journals/corr/abs-2508-02751,
  author       = {Yi Zhao and
                  Yajuan Peng and
                  Cam{-}Tu Nguyen and
                  Zuchao Li and
                  Xiaoliang Wang and
                  Hai Zhao and
                  Xiaoming Fu},
  title        = {SmallKV: Small Model Assisted Compensation of {KV} Cache Compression
                  for Efficient {LLM} Inference},
  journal      = {CoRR},
  volume       = {abs/2508.02751},
  year         = {2025},
  url          = {https://doi.org/10.48550/arXiv.2508.02751},
  doi          = {10.48550/ARXIV.2508.02751},
  eprinttype    = {arXiv},
  eprint       = {2508.02751},
  bibsource    = {dblp computer science bibliography, https://dblp.org}
}

@inproceedings{zhao-etal-2025-dac,
    title = "{DAC}: A Dynamic Attention-aware Approach for Task-Agnostic Prompt Compression",
    author = "Zhao, Yi  and
      Li, Zuchao  and
      Zhao, Hai  and
      Qi, Baoyuan  and
      Guoming, Liu",
    editor = "Che, Wanxiang  and
      Nabende, Joyce  and
      Shutova, Ekaterina  and
      Pilehvar, Mohammad Taher",
    booktitle = "Proceedings of the 63rd Annual Meeting of the Association for Computational Linguistics (Volume 1: Long Papers)",
    month = jul,
    year = "2025",
    address = "Vienna, Austria",
    publisher = "Association for Computational Linguistics",
    url = "https://aclanthology.org/2025.acl-long.952/",
    doi = "10.18653/v1/2025.acl-long.952",
    pages = "19395--19407",
    ISBN = "979-8-89176-251-0"
}

@inproceedings{tang-etal-2025-spindlekv,
    title = "{S}pindle{KV}: A Novel {KV} Cache Reduction Method Balancing Both Shallow and Deep Layers",
    author = "Tang, Zicong  and
      Luohe, Shi  and
      Li, Zuchao  and
      Qi, Baoyuan  and
      Guoming, Liu  and
      Zhang, Lefei  and
      Wang, Ping",
    editor = "Che, Wanxiang  and
      Nabende, Joyce  and
      Shutova, Ekaterina  and
      Pilehvar, Mohammad Taher",
    booktitle = "Proceedings of the 63rd Annual Meeting of the Association for Computational Linguistics (Volume 1: Long Papers)",
    month = jul,
    year = "2025",
    address = "Vienna, Austria",
    publisher = "Association for Computational Linguistics",
    url = "https://aclanthology.org/2025.acl-long.1380/",
    doi = "10.18653/v1/2025.acl-long.1380",
    pages = "28428--28442",
    ISBN = "979-8-89176-251-0"
}

\newpage
\makeatletter
\@ifundefined{isChecklistMainFile}{
  % We are compiling a standalone document
  \newif\ifreproStandalone
  \reproStandalonetrue
}{
  % We are being \input into the main paper
  \newif\ifreproStandalone
  \reproStandalonefalse
}
\makeatother

\ifreproStandalone
\documentclass[letterpaper]{article}
\usepackage[submission]{aaai2026}
\setlength{\pdfpagewidth}{8.5in}
\setlength{\pdfpageheight}{11in}
\usepackage{times}
\usepackage{helvet}
\usepackage{courier}
\usepackage{xcolor}
\frenchspacing

\begin{document}
\fi
\setlength{\leftmargini}{20pt}
\makeatletter\def\@listi{\leftmargin\leftmargini \topsep .5em \parsep .5em \itemsep .5em}
\def\@listii{\leftmargin\leftmarginii \labelwidth\leftmarginii \advance\labelwidth-\labelsep \topsep .4em \parsep .4em \itemsep .4em}
\def\@listiii{\leftmargin\leftmarginiii \labelwidth\leftmarginiii \advance\labelwidth-\labelsep \topsep .4em \parsep .4em \itemsep .4em}\makeatother

\setcounter{secnumdepth}{0}
\renewcommand\thesubsection{\arabic{subsection}}
\renewcommand\labelenumi{\thesubsection.\arabic{enumi}}

\newcounter{checksubsection}
\newcounter{checkitem}[checksubsection]

\newcommand{\checksubsection}[1]{%
  \refstepcounter{checksubsection}%
  \paragraph{\arabic{checksubsection}. #1}%
  \setcounter{checkitem}{0}%
}

\newcommand{\checkitem}{%
  \refstepcounter{checkitem}%
  \item[\arabic{checksubsection}.\arabic{checkitem}.]%
}
\newcommand{\question}[2]{\normalcolor\checkitem #1 #2 \color{blue}}
\newcommand{\ifyespoints}[1]{\makebox[0pt][l]{\hspace{-15pt}\normalcolor #1}}

\section*{Reproducibility Checklist}

\vspace{1em}
\hrule
\vspace{1em}

\textbf{Instructions for Authors:}

This document outlines key aspects for assessing reproducibility. Please provide your input by editing this \texttt{.tex} file directly.

For each question (that applies), replace the ``Type your response here'' text with your answer.

\vspace{1em}
\noindent
\textbf{Example:} If a question appears as
\begin{center}
\noindent
\begin{minipage}{.9\linewidth}
\ttfamily\raggedright
\string\question \{Proofs of all novel claims are included\} \{(yes/partial/no)\} \\
Type your response here
\end{minipage}
\end{center}
you would change it to:
\begin{center}
\noindent
\begin{minipage}{.9\linewidth}
\ttfamily\raggedright
\string\question \{Proofs of all novel claims are included\} \{(yes/partial/no)\} \\
yes
\end{minipage}
\end{center}
Please make sure to:
\begin{itemize}\setlength{\itemsep}{.1em}
\item Replace ONLY the ``Type your response here'' text and nothing else.
\item Use one of the options listed for that question (e.g., \textbf{yes}, \textbf{no}, \textbf{partial}, or \textbf{NA}).
\item \textbf{Not} modify any other part of the \texttt{\string\question} command or any other lines in this document.\\
\end{itemize}

You can \texttt{\string\input} this .tex file right before \texttt{\string\end\{document\}} of your main file or compile it as a stand-alone document. Check the instructions on your conference's website to see if you will be asked to provide this checklist with your paper or separately.

\vspace{1em}
\hrule
\vspace{1em}

% The questions start here

\checksubsection{General Paper Structure}
\begin{itemize}

\question{Includes a conceptual outline and/or pseudocode description of AI methods introduced}{(yes/partial/no/NA)}
yes

\question{Clearly delineates statements that are opinions, hypothesis, and speculation from objective facts and results}{(yes/no)}
yes

\question{Provides well-marked pedagogical references for less-familiar readers to gain background necessary to replicate the paper}{(yes/no)}
yes

\end{itemize}
\checksubsection{Theoretical Contributions}
\begin{itemize}

\question{Does this paper make theoretical contributions?}{(yes/no)}
no

	\ifyespoints{\vspace{1.2em}If yes, please address the following points:}
        \begin{itemize}
	
	\question{All assumptions and restrictions are stated clearly and formally}{(yes/partial/no)}
	Type your response here

	\question{All novel claims are stated formally (e.g., in theorem statements)}{(yes/partial/no)}
	Type your response here

	\question{Proofs of all novel claims are included}{(yes/partial/no)}
	Type your response here

	\question{Proof sketches or intuitions are given for complex and/or novel results}{(yes/partial/no)}
	Type your response here

	\question{Appropriate citations to theoretical tools used are given}{(yes/partial/no)}
	Type your response here

	\question{All theoretical claims are demonstrated empirically to hold}{(yes/partial/no/NA)}
	Type your response here

	\question{All experimental code used to eliminate or disprove claims is included}{(yes/no/NA)}
	Type your response here
	
	\end{itemize}
\end{itemize}

\checksubsection{Dataset Usage}
\begin{itemize}

\question{Does this paper rely on one or more datasets?}{(yes/no)}
yes

\ifyespoints{If yes, please address the following points:}
\begin{itemize}

	\question{A motivation is given for why the experiments are conducted on the selected datasets}{(yes/partial/no/NA)}
	yes

	\question{All novel datasets introduced in this paper are included in a data appendix}{(yes/partial/no/NA)}
	NA

	\question{All novel datasets introduced in this paper will be made publicly available upon publication of the paper with a license that allows free usage for research purposes}{(yes/partial/no/NA)}
	NA

	\question{All datasets drawn from the existing literature (potentially including authors' own previously published work) are accompanied by appropriate citations}{(yes/no/NA)}
	NA

	\question{All datasets drawn from the existing literature (potentially including authors' own previously published work) are publicly available}{(yes/partial/no/NA)}
	yes

	\question{All datasets that are not publicly available are described in detail, with explanation why publicly available alternatives are not scientifically satisficing}{(yes/partial/no/NA)}
	NA

\end{itemize}
\end{itemize}

\checksubsection{Computational Experiments}
\begin{itemize}

\question{Does this paper include computational experiments?}{(yes/no)}
yes

\ifyespoints{If yes, please address the following points:}
\begin{itemize}

	\question{This paper states the number and range of values tried per (hyper-) parameter during development of the paper, along with the criterion used for selecting the final parameter setting}{(yes/partial/no/NA)}
	yes

	\question{Any code required for pre-processing data is included in the appendix}{(yes/partial/no)}
	yes

	\question{All source code required for conducting and analyzing the experiments is included in a code appendix}{(yes/partial/no)}
	yes

	\question{All source code required for conducting and analyzing the experiments will be made publicly available upon publication of the paper with a license that allows free usage for research purposes}{(yes/partial/no)}
	yes
        
	\question{All source code implementing new methods have comments detailing the implementation, with references to the paper where each step comes from}{(yes/partial/no)}
	yes

	\question{If an algorithm depends on randomness, then the method used for setting seeds is described in a way sufficient to allow replication of results}{(yes/partial/no/NA)}
	yes

	\question{This paper specifies the computing infrastructure used for running experiments (hardware and software), including GPU/CPU models; amount of memory; operating system; names and versions of relevant software libraries and frameworks}{(yes/partial/no)}
	yes

	\question{This paper formally describes evaluation metrics used and explains the motivation for choosing these metrics}{(yes/partial/no)}
	yes

	\question{This paper states the number of algorithm runs used to compute each reported result}{(yes/no)}
	yes

	\question{Analysis of experiments goes beyond single-dimensional summaries of performance (e.g., average; median) to include measures of variation, confidence, or other distributional information}{(yes/no)}
	no

	\question{The significance of any improvement or decrease in performance is judged using appropriate statistical tests (e.g., Wilcoxon signed-rank)}{(yes/partial/no)}
	yes

	\question{This paper lists all final (hyper-)parameters used for each model/algorithm in the paper’s experiments}{(yes/partial/no/NA)}
	yes

\end{itemize}
\end{itemize}
\ifreproStandalone
\end{document}
\fi

\newpage

\appendix

\section{Training Hyperpameters}
\label{sec:appendix}

Reference Table~\ref{tab:hyperparameters}. Training takes about 72, 50 and 40 GPU-hours for 14B, 8B and 4B variants, with about 16, 12, 10 GPU-hours for data preparing in self-distillation.

\begin{table}[htbp]
\centering
\begin{tabular}{l|rr}
    \toprule
    Hyperparameter & \multicolumn{2}{c}{Value} \\
    \midrule
    Batch Size & \multicolumn{2}{c}{$2$} \\
    Grad. Acc. & \multicolumn{2}{c}{$8$} \\
    Max Seq. Len. & \multicolumn{2}{c}{$4096$} \\
    Num Epochs & \multicolumn{2}{c}{$2$} \\
    Total Steps & \multicolumn{2}{c}{$463,888$} \\
    \midrule
    \multirow{3}{*}{Max Learning Rate}
        & 5e-4 & (4B Model) \\
        & 3e-4 & (8B Model) \\
        & 2e-4 & (14B Model) \\
    \multirow{3}{*}{Min Learning Rate}
        & 1e-5 & (4B Model) \\
        & 1e-5 & (8B Model) \\
        & 1e-5 & (14B Model) \\
    Warm Up   & \multicolumn{2}{c}{5\% Total steps} \\
    Scheduler & \multicolumn{2}{c}{Cosine Annealing} \\
    \midrule
    Optimizer & \multicolumn{2}{c}{AdamW} \\
    Adam $\epsilon$ & \multicolumn{2}{c}{2e-4} \\
    Adam $\beta$s & \multicolumn{2}{c}{$(0.9,\,0.999)$} \\
    Weight Decay  & \multicolumn{2}{c}{$0.01$} \\
    \bottomrule 
\end{tabular}
\caption{Hyperparameters used for training.}
\label{tab:hyperparameters}
\end{table}

\section{Hardware Detail}
\label{sec:hdw}

Please reference Table~\ref{tab:hdw}.

\begin{table}[ht]
    \centering
    \begin{tabular}{c|p{5cm}}
    \toprule
        Item & Value \\
    \midrule
        CPU & 24 * Intel(R) Xeon(R) Silver 4314 CPU @ 2.40GHz \\
        GPU &  NVIDIA A800 PCIe 80 GB \\
        RAM & 212GB DDR4-2667 \\
    \bottomrule
    \end{tabular}
    \caption{Hardware used.}
    \label{tab:hdw}
\end{table}


\end{document}